\documentclass[lettersize,journal]{IEEEtran}
\usepackage{amsmath,amsfonts}
\usepackage{algorithmic}
\usepackage{algorithm}
\usepackage{array}
\usepackage{enumitem}
\usepackage[caption=false,font=normalsize,labelfont=sf,textfont=sf]{subfig}
\usepackage{textcomp}
\usepackage{stfloats}
\usepackage{url}
\usepackage{verbatim}
\usepackage{graphicx}
\usepackage{cite}
\usepackage{xcolor}
\usepackage{booktabs}
\usepackage{longtable}
\usepackage{multirow}
\usepackage{hyperref}
\usepackage{amsmath}
\usepackage{amssymb}

\begin{document}

\title{Machine Learning for Synthetic Data Generation: \\ A Review}


\author{Yingzhou Lu\IEEEauthorrefmark{4}, 
Lulu Chen\IEEEauthorrefmark{8},
Yuanyuan Zhang\IEEEauthorrefmark{9},
 Minjie Shen\IEEEauthorrefmark{2}, Huazheng Wang\IEEEauthorrefmark{3}, 
 Xiao Wang\IEEEauthorrefmark{6}, 
 Capucine van Rechem\IEEEauthorrefmark{4}, Tianfan Fu\IEEEauthorrefmark{1}, Wenqi Wei\IEEEauthorrefmark{7}\IEEEauthorrefmark{5}

%\IEEEauthorrefmark{5}
 
 % 
 % and Tianfan Fu\IEEEauthorrefmark{7}
        %and~Jane~Doe,~\IEEEmembership{Life~Fellow,~IEEE}% <-this % stops a space
%\thanks{M. Shell was with the Department of Electrical and Computer Engineering, Georgia Institute of Technology, Atlanta, GA, 30332 USA e-mail: (see http://www.michaelshell.org/contact.html).}% <-this % stops a space
\IEEEcompsocitemizethanks{
%\IEEEcompsocthanksitem Work done while the authors were at Georgia Institute of Technology.
% Wenqi Wei is with Computer and Information Sciences department, Fordham University, New York City, NY, 10019. \protect \\ 
%\IEEEcompsocthanksitem \indent Wenqi Wei, Ling Liu, Ka-Ho Chow, and Yanzhao Wu are with
\IEEEauthorrefmark{4}Department of Pathology, Stanford University, Stanford, CA, 94305. \\
\IEEEauthorrefmark{8}Department of Electrical and Computer Engineering, Virginia Polytechnic Institute and State University, Arlington, VA 22203, USA.  \\
\IEEEauthorrefmark{9}Department of Computer Science of Purdue University, West Lafayette, IN 47907, USA.  \\
%\IEEEcompsocthanksitem \indent Wenqi Wei is now with 
%Work partially done when the author was a PhD student at Georgia Institute of Technology\IEEEauthorrefmark{2}.
\IEEEauthorrefmark{2}The Bradley Department of Electrical and Computer Engineering, Virginia Tech \\
\IEEEauthorrefmark{3}School of Electrical Engineering and Computer Science, Oregon State University, 
Corvallis, OR, 97331. \\
%\IEEEcompsocthanksitem \indent Jingya Zhou is with 
\IEEEauthorrefmark{6}School of Computer Science \& Engineering, University of Washington, Seattle, WA, 98105. \\
\IEEEauthorrefmark{1}Computer Science Department, Rensselaer Polytechnic Institute, Troy, NY, 12180.  \\
\IEEEauthorrefmark{7}Computer and Information Science Department, Fordham University, New York City, NY, 10023.  \\
%\protect \\ 
\IEEEauthorrefmark{5}Corresponding author. 
% \indent Yanzhao Wu is with School of Computing and Information Sciences, Florida International University, Miami, FL, 33199. \protect\\ 
 \protect\\ %Work done while the author was visiting Georgia Institute of Technology.
%  \indent Mehmet Emre Gursoy is with Department of Computer Engineering, Koc University, Istanbul, Turkey, 34450. \protect\\ 
% \indent Stacey Truex is with Department of Computer Science, Denison University, Granville, Ohio, 43023. \protect\\ 
 E-mails: %\{wenqiwei,yanzhaowu,khchow\}@gatech.edu,
%cvrechem@stanford.edu, 
% fut2@rpi.edu, 
%huazheng.wang@oregonstate.edu, 
%cvrechem@stanford.edu, 
wenqiwei@fordham.edu.} % <-this % stops an unwanted space 
%\hfil\break # for a new line
\thanks{Manuscript received xxxx xx, xxxx; revised xxxxx xx, xxxx.}
}

\markboth{Journal of \LaTeX\ Class Files,~Vol.~14, No.~8, August~2021}%
{Shell \MakeLowercase{\textit{et al.}}: A Sample Article Using IEEEtran.cls for IEEE Journals}


\maketitle

\begin{abstract}
% Data plays a pivotal role in machine learning. However, in real-world applications, several issues arise with data, such as poor quality, limited data points leading to the under-fitting of machine learning models, and difficulties in accessing data due to privacy, safety, and regulatory concerns. \textit{Synthetic data generation} offers a promising alternative, as it enables data sharing and usage in ways that real-world data cannot. This paper presents a systematic review of existing works that leverage machine learning models for synthetic data generation. Specifically, we discuss synthetic data generation from several perspectives: (i) applications, encompassing computer vision, speech, natural language, healthcare, and business; (ii) machine learning methods, with a focus on neural network architectures and deep generative models; (iii) privacy and fairness concerns.
% Moreover, we identify the challenges and opportunities in this emerging field and propose future research directions.
Machine learning heavily relies on data, but real-world applications often encounter various data-related issues. These include data of poor quality, insufficient data points leading to under-fitting of machine learning models, and difficulties in data access due to concerns surrounding privacy, safety, and regulations. In light of these challenges, the concept of \textit{synthetic data generation} emerges as a promising alternative that allows for data sharing and utilization in ways that real-world data cannot facilitate. This paper presents a comprehensive systematic review of existing studies that employ machine learning models for the purpose of generating synthetic data. The review encompasses various perspectives, starting with the applications of synthetic data generation, spanning computer vision, speech, natural language processing, healthcare, and business domains. Additionally, it explores different machine learning methods, with particular emphasis on neural network architectures and deep generative models. The paper also addresses the crucial aspects of privacy and fairness concerns related to synthetic data generation. Furthermore, this study identifies the challenges and opportunities prevalent in this emerging field, shedding light on the potential avenues for future research. By delving into the intricacies of synthetic data generation, this paper aims to contribute to the advancement of knowledge and inspire further exploration in synthetic data generation.
\end{abstract}


\section{Introduction}

\IEEEPARstart{M}{achine} learning endows intelligent computer systems with the capacity to autonomously tackle tasks, pushing the envelope of industrial innovation~\cite{ng2016artificial}. By integrating high-performance computing, contemporary modeling, and simulations, machine learning has evolved into an indispensable instrument for managing and analyzing massive volumes of data~\cite{boden1996artificial,haenlein2019brief}.

Nonetheless, it is important to recognize that machine learning does not invariably resolve problems or yield the optimal solution. Despite artificial intelligence is currently experiencing a golden age, numerous challenges persist in the development and application of machine learning technology~\cite{lucini2021real}. As the field continues to advance, addressing these obstacles will be essential for unlocking the full potential of machine learning and its transformative impact on various industries.

%%%%%%%%%%%%% these information may be useful, provided for your reference
% Aside from the scientific literature, data augmentation is a topic covered by commercial
% tools, such as Synthesized (https://www.synthesized.io/), which provides a Python library
% measuring the extent of bias in a dataset, together with reporting facilities, the capability to
% flag sensitive variables and fairness scoring tools. Clearbox AI (https://app.clearbox.ai)
% is a data augmentation tool performing anonymization and reducing overfitting and
% imbalances starting from an original dataset. DataGen (https://www.ceadar.ie/pages/
% datagen/) offers the capability to generate datasets according to a set of features defined
% by the user; however, when the process is automatically executed, it must be based on an
% existing dataset. Manual generation is made possible as the user can specify in advance a
% set of relationships among the dataset variables.
%%%%%%%%%%%%%%%
The process of collecting and annotating data is both time-consuming and expensive~\cite{jordan2015machine}, giving rise to numerous issues. As machine learning is heavily dependent on data, some of the key hurdles and challenges it faces include: 
\begin{itemize}[leftmargin=*]
\item \textbf{Data quality}.
Ensuring data quality is one of the most significant challenges confronting machine learning professionals. When data is of subpar quality, models may generate incorrect or imprecise predictions due to confusion and misinterpretation~\cite{pipino2002data}~\cite{shen2022comparative}.
\item \textbf{Data scarcity}.
A considerable portion of the contemporary AI dilemma stems from inadequate data availability: either the number of accessible datasets is insufficient, or manual labeling is excessively costly~\cite{babbar2019data}.
\item \textbf{Data privacy and fairness}.
There are many areas in which datasets cannot be publicly released due to privacy ad fair issues. In these cases,
generating synthetic data can be very useful, and we will investigate ways of creating anonymized datasets with
differential privacy protections.
\end{itemize}

\begin{figure*}
\centering
\includegraphics[width=0.9\linewidth]{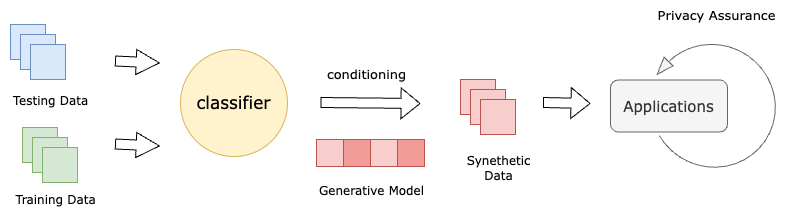}
\caption{Synthetic data generation. }
\label{fig:framework}
\end{figure*}

Tackling these issues is crucial to fully realizing the transformative power of machine learning across diverse sectors~\cite{nikolenko2021synthetic,bolon2013review,frid2018synthetic}. Generally, synthetic data are defined as the artificially annotated information generated by computer algorithms or simulations~\cite{wang2019learning,lucini2021real}. 
In many cases, synthetic data is necessary when real data is either unavailable or must be kept private due to privacy or compliance risks~\cite{bolon2013review,abowd2008protective,abay2019privacy}. This technology is extensively utilized in various sectors, such as healthcare, business, manufacturing, and agriculture, with demand growing at an exponential rate~\cite{raghunathan2021synthetic}.

The objective of this paper is to offer a high-level overview of several state-of-the-art approaches currently being investigated by machine learning researchers for synthetic data generation. For the reader's convenience, we summarize the paper's main contributions as follows:

\begin{itemize}
\item We present pertinent ideas and background information on synthetic data, serving as a guide for researchers interested in this domain.
\item We explore different real-world application domains and emphasize the range of opportunities that GANs and synthetic data generation can provide in bridging gaps (Section~\ref{sec:application}).
\item We examine a diverse array of deep generative models dedicated to generating high-quality synthetic data, present advanced generative models, and outline potential avenues for future research (Section~\ref{sec:generative_ai}).
\item We address privacy and fairness concerns, as sensitive information can be inferred from synthesized data, and biases embedded in real-world data can be inherited. We review current technological advancements and their limitations in safeguarding data privacy and ensuring the fairness of synthesized data (Section~\ref{sec:privacy} and~\ref{sec:fairness}).
\item We outline several general evaluation strategies to assess the quality of synthetic data (Section~\ref{sec:evaluation}).
\item We identify challenges faced in generating synthetic data and during the deployment process, highlighting potential future work that could further enhance functionality (Section~\ref{sec:challenge}).
\end{itemize}

% This paper is organized as follows. 
% The first section provides an in-depth overview of the topic. Section two includes a collection of applications of synthetic data generation.
% Models that have made a significant impact on the filed are presented in the section three. In the next three sections, we will go over the progress and issues to be solved. The last section covers a summary of this topic, highlighting the potential future work that could further enhance the functionality.
% We also list and compare all the representative synthetic data generation methods in Table~\ref{table:summarize}.  

% from business view: \url{https://research.aimultiple.com/synthetic-data-generation/}

% \begin{longtable}{|p{0.24\columnwidth}|p{0.75\columnwidth}|}
% \caption{Glossary of terms.
% \label{table:terminology}}
% \\\toprule
% \textbf{Term} & \textbf{Description}  
% \\\midrule 
% quantum computation &  \\ \hline 
% NISQ & \\ 
% Machine Learning & \\
% deep learning & \\ 
% neural network & \\ 
% quantum computation & \\ 
% \\\bottomrule[0.6pt]
% \end{longtable}

\begin{table*}[ht!]
\centering
\caption{Summarization of representative works in synthetic data generation.}
%\resizebox{\columnwidth}{!}{
\scalebox{0.99}{
\begin{tabular}{p{0.25\columnwidth}|p{0.23\columnwidth}|p{0.40\columnwidth}|p{0.38\columnwidth}|p{0.48\columnwidth}}
\toprule[0.9pt]
Paper & Application & Generative AI & DNN & Dataset \\ 
\hline 
MedGAN~\cite{choi2017generating} & healthcare & GAN & MLP & MIMIC/Sutter (Electronic health record) \\ 
MMCGAN~\cite{ziegler2022multi} & healthcare \& CV & GAN & CNN &  chest CT images \\ 
DeepSynth~\cite{dunn2019deepsynth} & healthcare \& CV & GAN & CNN & rat kidney tissue (microscope image)\\ 
ChemSpaceE~\cite{du2022chemspace} & drug & VAE & GNN & ZINC (drug molecule)~\cite{sterling2015zinc} \\ 
JTVAE~\cite{jin2018junction} & drug & VAE & GNN & ZINC (drug molecules)~\cite{sterling2015zinc} \\ 
REINVENT~\cite{olivecrona2017molecular} & drug & RL & RNN & ZINC (drug molecules)~\cite{sterling2015zinc} \\ 
CORE~\cite{fu2020core} & drug & VAE & GNN & ZINC (drug molecule)~\cite{sterling2015zinc} \\ 
RGA~\cite{fu2022reinforced} & drug & RL & geometric NN & ZINC and TDC~\cite{huang2022artificial} \\ 
CorGAN~\cite{torfi2020corgan} & healthcare & GAN & CNN & MIMIC-III dataset, UCI Epileptic Seizure Recognition
dataset  \\ 
DAAE~\cite{lee2020generating} & healthcare & VAE+GAN & recurrent autoencoder & MIMIC-III, UT Physicians clinical databases\\ 
HAPNEST~\cite{wharrie2022hapnest}& healthcare & approximate Bayesian computation (not deep learning) & NA (w.o. DNN) & Genomes Project and HGDP datasets \\ 
synthpop~\cite{nowok2016synthpop} & healthcare & proper synthesis & Statistical hypothesis testing
 & SD2011 \\
CycleGAN~\cite{zhu2017unpaired} & vision & GAN & CNN & pix2pix  \\ 
DP-CGAN~\cite{torkzadehmahani2019dp}& vision & GAN & deep
CGAN & MNIST  \\ 
BigGANs~\cite{brock2018large}&  vision & GAN & large scale GAN & ImageNet \\ 
VideoDiff~\cite{ho2022video} & vision & diffusion & CNN & BAIR Robot Pushing,  Kinetics-600  \\ 
VQ-VAE~\cite{razavi2019generating} & vision & VAE & PixelCNN & ImageNet \\ 
GIRAFFE~\cite{niemeyer2021giraffe} & vision & GAN & CNN & CompCars, LSUN Churches, and FFHQ \\ 
Wavegrad~\cite{chen2020wavegrad} & TTS & diffusion & gradient-based sampling &  LJ Speech \\ 
TTS-GAN~\cite{guo2019new} & TTS & GAN & auto-regressive model & Tacotron2  \\ 
Seq-GAN~\cite{yu2017seqgan} & NLP & GAN+RL & CNN & Nottingham dataset \\ 
BLEURT~\cite{sellam2020bleurt}  & NLP  & Language model & BERT & WebNLG
Competition dataset \\ 
TextGen-RL~\cite{shi2018toward} & NLP & RL & LSTM &  \\ 
SynBench~\cite{ko2022synbench} & NLP & conditional Gaussian mixture & & CIFAR10 \\ 
RelGAN~\cite{nie2018relgan} & image and text & GAN & CNN & COCO Image Captions
dataset\\ 
DPGM~\cite{acs2018differentially} & audio and text & generative artificial neural networks & differentially private kernel k-means & MNIST, anonymized Call
Detail Record (CDR) \\ 

WaveGAN~\cite{donahue2018adversarial}& audio & GAN & DCGAN & Speech Commands Dataset \\ 
Wavenet~\cite{oord2016wavenet}& audio & GAN & LSTM & CSTR voice corpula (multi-channel English audio) \\ 
Stutter-TTS~\cite{zhang2022stutter} & audio & 
phonetic encoder and the
decoder & CNN & recordings \\ 
Quant GANs~\cite{wiese2020quant} & business & GAN & MLP+ Temporal convolutional networks(TCN) & simulated data \\ 
CGAN~\cite{fu2019time}  & business & GAN & CNN & Vector autoregressive (VAR) time series\\ 
PATE-GAN~\cite{jordon2018pate} & business & GAN & Private Aggregation of Teacher Ensembles (PATE) & Kaggle \\
CollGAN~\cite{atlas2022deep} & physics (particle collision) & VAE/GAN & MLP & ATLAS \\
\bottomrule[0.9pt]
%%% you may add this paper Synthetic data augmentation using GAN for improved liver lesion classification.
\end{tabular}
}
\label{table:summarize}
\end{table*}

\section{Application}
\label{sec:application}
Synthetic data offers a multitude of compelling advantages, making it a highly appealing option for a wide range of applications. By streamlining the processes of training, testing, and deploying AI solutions, synthetic data facilitates more efficient and effective development. Furthermore, this cutting-edge technology reduces the risk of exposing sensitive information, thereby ensuring customer security and privacy~\cite{lucini2021real}.

As researchers transition synthetic data from the lab to practical implementations, its real-world applications continue to broaden. This section explores several notable domains where synthetic data generation substantially impacts addressing real-world challenges.

\subsection{Vision}
Supervised learning relies heavily on the availability of labeled data~\cite{dewi2022synthetic}. However, in many applications, particularly in computer vision, manual labeling is often necessary~\cite{zhao2021mt,yi2018enhance}. Tasks such as segmentation, depth estimation, and optical flow estimation can be exceedingly challenging to label manually. Synthetic data has emerged as a transformative solution in this context, significantly improving the labeling process~\cite{chen2019learning}.

Sankaranarayanan et al. introduced a generative adversarial network (GAN) that narrows the gap between embeddings in the learned feature space, facilitating Visual Domain Adaptation~\cite{sankaranarayanan2018learning}. This approach enables semantic segmentation across different domains. The GAN uses a generator to project features onto the image space, which the discriminator subsequently operates on. Adversarial losses can be derived from the discriminator's output~\cite{dong2019towards}. Notably, applying adversarial losses to the projected image space has been shown to yield significantly better performance compared to applying them directly to the feature space~\cite{sankaranarayanan2018learning}.

In a recent study, a Microsoft research team demonstrated the effectiveness of synthetic data in face-related tasks by combining a parametric 3D face model with an extensive library of hand-crafted assets~\cite{wood2021fake}. This approach rendered training images with remarkable realism and diversity. The researchers trained machine learning systems for tasks such as landmark localization and face parsing using synthetic data, showing that it can achieve comparable accuracy to real data. Furthermore, synthetic data alone proved sufficient for detecting faces in unconstrained settings~\cite{wood2021fake}. 

\begin{figure}
\centering
\includegraphics[width=\linewidth]{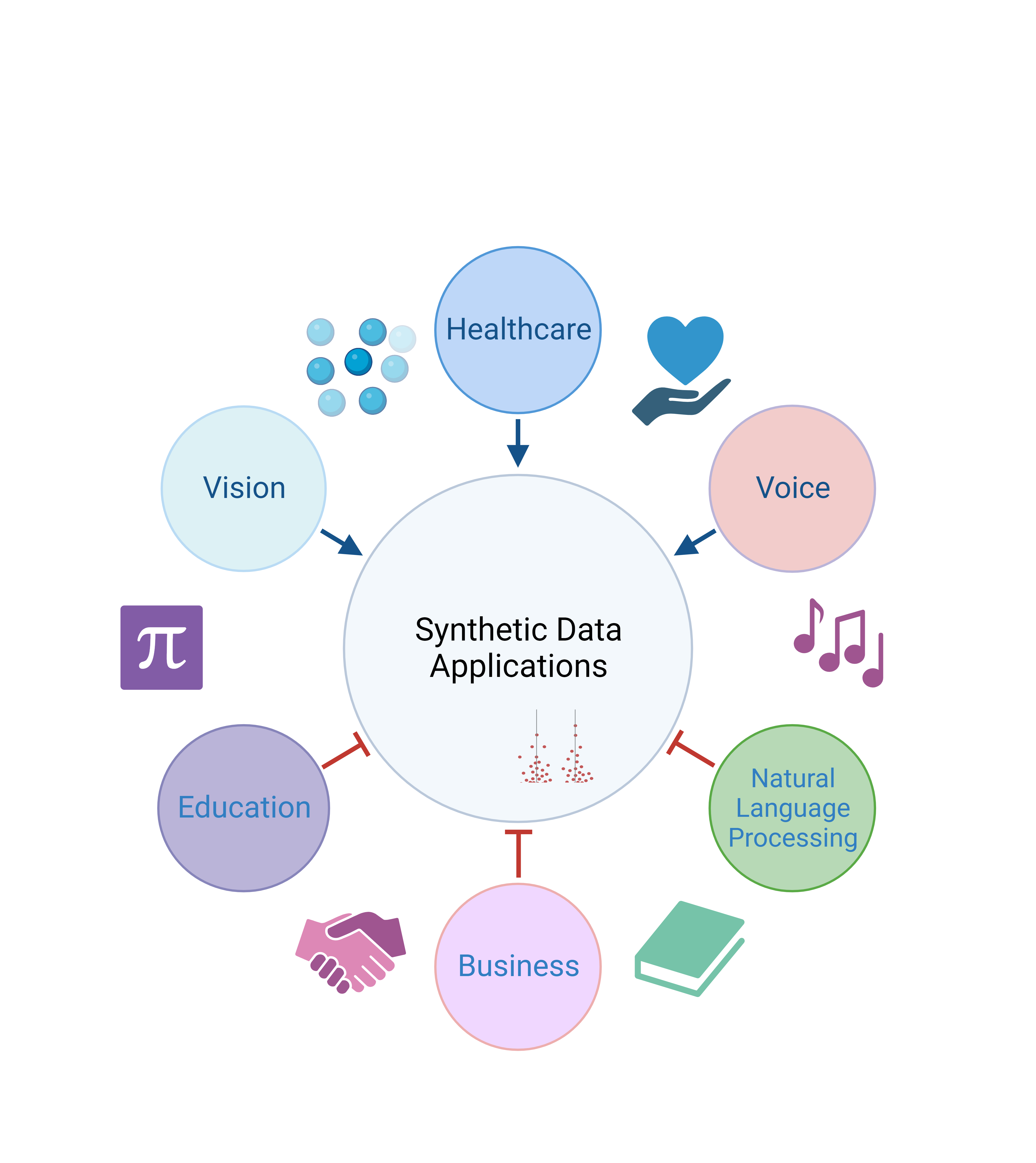}
\vspace{-1cm}
\caption{Synthetic data applications }
\label{fig:applications}
\end{figure}

\subsection{Voice}
The field of synthetic voice is at the forefront of technological advancement, and its evolution is happening at a breakneck pace. With the advent of machine learning and deep learning, creating synthetic voices for various applications such as video production, digital assistants, and video games~\cite{werchniak2021exploring} has become easier and more accurate. This field is an intersection of diverse disciplines, including acoustics, linguistics, and signal processing. Researchers in this area continuously strive to improve synthetic voices' accuracy and naturalness. As technology advances, we can expect to see synthetic voices become even more prevalent in our daily lives, assisting us in various ways and enriching our experiences in many fields~\cite{li2016feature}.

The earlier study includes spectral modeling for statistical parametric speech synthesis, in which low-level, un-transformed spectral envelope parameters are used for voice synthesis. The low-level spectral envelopes are represented by graphical models incorporating multiple hidden variables, such as restricted Boltzmann machines and deep belief networks (DBNs)~\cite{fu2014tandem}. The proposed conventional hidden Markov model (HMM)-based speech synthesis system can be significantly improved in terms of naturalness and over-smoothing~\cite{ling2013modeling}.

Synthetic data can also be applied to Text-to-Speech (TTS) to achieve near-human naturalness~\cite{fazel2021synthasr,li2015improved}. As an alternative to sparse or limited data, synthetic speech (SynthASR) was developed for automatic speech recognition. The combination of weighted multi-style training, data augmentation, encoder freezing, and parameter regularization is also employed to address catastrophic forgetting.
Using this novel model, the researchers were able to apply state-of-the-art techniques to train a wide range of end-to-end (E2E) automatic speech recognition (ASR) models while reducing the need for production data and the costs associated with it~\cite{fazel2021synthasr}.

\subsection{Natural Language Processing (NLP)}

The increasing interest in synthetic data has spurred the development of a wide array of deep generative models in the field of natural language processing (NLP)~\cite{dewi2022synthetic}. In recent years, a multitude of methods and models have illustrated the capabilities of machine learning in categorizing, routing, filtering, and searching for relevant information across various domains~\cite{forman2003extensive}.

Despite these advancements, challenges remain. For example, the meaning of words and phrases can change depending on their context, and homonyms with distinct definitions can pose additional difficulties~\cite{yue2022synthetic}. To tackle these challenges, the BLEURT model was proposed, which models human judgments using a limited number of potentially biased training examples based on BERT. The researchers employed millions of synthetic examples to develop an innovative pre-training scheme, bolstering the model's ability to generalize~\cite{zheng2021using,fu2024ddn3}. Experimental results indicate that BLEURT surpasses its counterparts on both the WebNLG Competition dataset and the WMT Metrics, highlighting its efficacy in NLP tasks~\cite{sellam2020bleurt}.

Another significant breakthrough in text generation using GANs is RelGAN, developed by Rice University. This model is comprised of three main components: a relational memory-based generator, a Gumbel-Softmax relaxation algorithm, and multiple embedded representations within the discriminator. When benchmarked against several cutting-edge models, RelGAN demonstrates superior performance in terms of sampling quality and diversity. This showcases its potential for further investigation and application in a wide range of NLP tasks and challenges~\cite{nie2018relgan,zhao2022act}.

\subsection{Healthcare}
In order to protect health information and improve reproducibility in research, synthetic data has drawn mainstream attention in the healthcare industry~\cite{chen2021synthetic,tucker2020generating}. 
Many labs and companies have harnessed the tools of big data and advanced computation tools to produce large quantities of synthetic data, or digital twin~\cite{wang2024twin,lu2022cot}. 
Modeled after patient data, synthetic data generation is essential to understanding diseases while maintaining patient confidentiality and privacy simultaneously~\cite{dahmen2019synsys}. 
Theoretically, synthetic data can reflect the original distribution of the data instead of revealing actual patient data~\cite{dahmen2019synsys,lu2019integrated,wang2021generating}.

Synthetic data generation can also be utilized to discover new scientific principles by grounding it in biological priors~\cite{chen2021synthetic}. 
There have been a good number of models and software developed, such as SynSys, which uses hidden Markov models and regression models initially trained on real datasets to generate synthetic time series data consisting of nested sequences~\cite{tucker2020generating}; 
and corGAN, in which synthetic data is generated by capturing correlations between adjacent medical features in the data representation space~\cite{torfi2020corgan}.

Synthetic data generation has also been widely used in drug discovery, especially \textit{de novo} drug molecular design. Drugs are essentially molecular structures with desirable pharmaceutical properties. The goal of \textit{de novo} drug design is to produce novel and desirable molecule structures from scratch. The word ``\textit{de novo}'' means from the beginning. 
The whole molecule space is around $10^{60}$~\cite{bohacek1996art,huang2022artificial,huang2020deeppurpose}. Most of the existing methods rely heavily on brute-force enumeration and are computationally prohibitive. 
Generative models are able to learn the distribution of drug molecules from the existing drug database and then draw novel samples (i.e., drug molecules) from the learned molecule distribution, including variational autoencoder (VAE)~\cite{gomez2018automatic,zhang2021ddn2,jin2018junction}, generative adversarial network (GAN)~\cite{de2018molgan}, energy-based model (EBM)~\cite{fu2022antibody,fu2021differentiable}, diffusion model~\cite{xu2021geodiff}, reinforcement learning (RL)~\cite{olivecrona2017molecular,zhou2019optimization,fu2022reinforced}, genetic algorithm~\cite{jensen2019graph}, sampling-based methods~\cite{fu2021mimosa,fu2022sipf}, etc.

In healthcare, patient information is often stored in electronic health records (EHR) format~\cite{kruse2017security,wen2022disentangled,fu2019pearl}. 
Research in medicine has been greatly facilitated by the availability of information from electronic health records~\cite{goncalves2020generation,du2023abds}. MedGAN, an adversarial network model for generating realistic synthetic patient records, has been proposed by Edward Choi and other colleagues. With the help of an autoencoder and generative adversarial networks, medGAN can generate high-dimensional discrete variables (e.g., binary and count features) based on real patient records~\cite{choi2017generating}. Based on their evaluations of medGAN's performance on a set of diverse tasks reported, including reporting distribution statistics, classification performance~\cite{lu2018multi}, and expert review, medGAN exhibits close-to-real-time performance~\cite{bhanot2021problem,fu2022hint,fu2019ddl,chen2021data,choi2017generating}. 
Using synthetic data can help reduce the regulatory barriers preventing the widespread sharing and integration of patient data across multiple organizations in the past~\cite{eigenschink2021deep,wu2022cosbin}. 
Researchers across the globe would be able to request access to synthetic data from an institution to conduct their own research using the data. Such capabilities can increase both the efficiency and scope of the study as well as reduce the likelihood of biases being introduced into the results~\cite{tucker2020generating,wang2022eeg,du2022molgensurvey}.

\subsection{Business}
\label{sec:business}

The inherent risk of compromising or exposing original data persists as long as it remains in use, particularly in the business sector, where data sharing is heavily constrained both within and outside the organization~\cite{el2020practical}. Consequently, it is crucial to explore methods for generating financial datasets that emulate the properties of "real data" while maintaining the privacy of the involved parties~\cite{el2020practical}.

Efforts have been made to secure original data using technologies like encryption, anonymization, and cutting-edge privacy preservation~\cite{mannino2019real}. However, information gleaned from the data may still be employed to trace individuals, thereby posing the risk~\cite{assefa2020generating}. A notable advantage of synthetic data lies in its ability to eliminate the exposure of critical data, thus ensuring privacy and security for both companies and their customers~\cite{lu2019empirical}. 
Moreover, synthetic data enables organizations to access data more rapidly, as it bypasses privacy and security protocols~\cite{hittmeir2019utility}. In the past, institutions possessing extensive data repositories could potentially assist decision-makers in resolving a broad spectrum of issues. However, accessing such data, even for internal purposes, was hindered by confidentiality concerns. Presently, companies are harnessing synthetic data to refresh and model original data, generating continuous insights that contribute to enhancing the organization's performance~\cite{lucini2021real}.

\subsection{Education}
Synthetic data is gaining increasing attention in the field of education due to its vast potential for research and teaching. Synthetic data refers to computer-generated information that mimics the properties of real-world data without disclosing any personally identifiable information~\cite{berg2016role}. This approach proves instrumental for educational settings, where ethical constraints often limit the use of real-world student data. Therefore, synthetic data offers a robust solution for privacy-concerned data sharing and analysis, enabling the creation of accurate models and strategies to improve the teaching-learning process.

A detailed example of synthetic data usage in education is the simulation of student performance data to aid in designing teaching strategies. Suppose an educational researcher wants to investigate the impact of teaching styles on student performance across different backgrounds and learning abilities. However, obtaining real student data for such studies can be ethically complex and potentially intrusive. In such a situation, synthetic data can be generated that mirrors the demographic distributions, learning patterns, and likely performance of a typical student population. This data can then be used to model the effects of various teaching strategies without compromising student privacy~\cite{howe2017synthetic}.

Furthermore, synthetic data can be a powerful tool in teacher training programs. For example, teacher candidates can use synthetic student data to practice data-driven instructional strategies, including differentiated instruction and personalized learning plans. They can analyze this synthetic data, identify patterns, determine student needs, and adjust their instructional plans accordingly. By using synthetic data, teacher candidates gain practical experience in analyzing student data and adapting their teaching without infringing on the privacy of actual students~\cite{bautista2021protecting}. Thus, synthetic data serves as a valuable bridge between theory and practice in education, driving innovation while safeguarding privacy.

\subsection{Location and Trajectory Generation}

Location and trajectory are a particular form of data that could highly reflect users’ daily lives, habits, home addresses, workplaces, etc. To protect location privacy, synthetic location generation is introduced as opposed to location perturbation~\cite{jiang2021location}. %andres2013geo,
The main challenge of generating synthetic location and trajectory data is to resemble genuine user-produced data while offering practical privacy protection simultaneously. One approach to generating the location and trajectory data is to inject a synthetic point-based site within a user’s trajectory~\cite{kato2012dummy,du2021graphgt}. 

Synthetic trajectory generation is frequently combined with privacy-enhancing techniques to further prevent sensitive inference from the synthesized data. 
For example, 
Chen et al.~\cite{chen2012differentially} introduces an N-gram-based method to predict the following position based on previous positions for publishing trajectory. They exploit the prefix tree to describe the n-gram model while combining it with differential privacy~\cite{dwork2014algorithmic}. 
\cite{cunningham2021real} extends the n-gram model with local differential privacy and \cite{du2023ldptrace} further replaces the n-gram model with key movement mobility for differentially private trajectory generation. 
By comparison, 
\cite{he2015dpt} proposes a synthetic trajectory strategy based on the discretization of raw trajectories using hierarchical reference systems to capture individual movements at differing speeds. Their method adaptively selects a small set of reference systems and constructs prefix tree counts with differential privacy. Applying direction-weighted sampling, 
the decrease in tree nodes reduces the amount of added noise and improves the utility of the synthetic data. \cite{wang2017protecting} constructs the differentially private prefix tree and calibrates original trajectories against a selection of anchor points. By extracting multiple differential private distributions with redundant information~\cite{gursoy2018utility,gursoy2018differentially}, the authors generate a new trajectory with samples from these distributions. By comparison, \cite{mir2013dp} estimates various distributions of an attribute set to determine trajectories and~\cite{roy2016practical} consider the interactions between different attributes by grouping strongly correlated attributes into non-disjoint sets and constructing a corresponding distribution for each set.

In addition to differential privacy, Bindschaedler and Shokri~\cite{bindschaedler2016synthesizing} enforce plausible deniability to generate privacy-preserving synthetic traces. It first introduces trace similarity and intersection functions
that map a fake trace to a real hint under similarity and intersection constraints. Then, it generates one fake trace by clustering the locations and replacing the trajectory locations with those from the same group. If the fake trace satisfies plausible deniability, i.e., there exist k other real traces that can map to the fake trace, then
it preserves the privacy of the seed trace. While existing studies mainly use the Markov chain model, \cite{wang2023privtrace} proposes PrivTrace, which 
%to choose between the first-order and second-order Markov model adaptively. The proposed PrivTrace 
controls the space and time overhead by the first-order Markov chain model and achieves good accuracy for next-step prediction by the second-order Markov chain model. 
\cite{narita2024synthesizing} considers the location synthesizer that generates location traces, including co-locations of friends, while offering node-level differential privacy for the friendship and user-level differential privacy for the co-location count matrix.

\subsection{AI-Generated Content (AIGC)}
\label{sec:aigc}

AI-Generated Content (AIGC) stands at the forefront of the technology and content creation industry, changing the dynamics of content production. A typical example of AIGC is OpenAI's ChatGPT, an AI-driven platform generating human-like text in response to prompts or questions. It leverages a vast corpus of internet text to generate detailed responses, often indistinguishable from those a human writer would produce. This capacity extends beyond simple question-answer pairs to crafting whole articles, stories, or technical explanations on a wide range of topics, thus creating a novel way of producing blog posts, articles, social media content, etc~\cite{cao2023comprehensive,vaswani2017attention}.

Google's Project Bard focuses more on the creative aspects of text generation. It is designed to generate interactive fiction and assist in storytelling. Users can engage in an interactive dialogue with the model, directing the course of a narrative by providing prompts that the AI responds to, thus co-creating a story. This opens up fascinating possibilities for interactive entertainment and digital storytelling~\cite{tao2023boundary}.

An innovative application of AIGC is in the field of news reporting. News agencies increasingly use AI systems, such as the GPT series, to generate news content. For instance, the Associated Press uses AI to generate news articles about corporate earnings automatically. The AI takes structured data about company earnings and transforms it into a brief, coherent, and accurate news report. This automation allows the agency to cover many companies that would be possible with human journalists alone~\cite{ventayen2023openai}.

Additionally, AIGC has found its place in the creative domain, with AI systems being used to generate book descriptions, plot outlines, and even full chapters of novels. For instance, a novelist could use ChatGPT to generate a synopsis for their upcoming book based on a few keywords or prompts related to the story. Similarly, marketing teams utilize AI to create compelling product descriptions for online marketplaces~\cite{yue2023democratizing}. This increases efficiency and provides a level of uniformity and scalability that would be challenging to achieve with human writers alone. %Through these examples, it is clear that 
Looking forward, 
AIGC is profoundly impacting the landscape of content creation and will continue to shape it in the future~\cite{tao2023boundary}.

\subsection{Finance}
Synthetic data generation offers significant benefits for the finance industry~\cite{assefa2020generating}, as detailed below. 
First, financial data is highly sensitive and subject to stringent privacy regulations~\cite{cheng2021can}. Synthetic data mimics real data without exposing actual customer information, enabling institutions to comply with privacy laws while still utilizing detailed datasets for analysis and development.
Second, synthetic data can be used to test and validate financial algorithms and models under various conditions. For example, trading algorithms can be tested using synthetic market data to evaluate their performance under different market scenarios, including rare or extreme events that may not be present in historical data~\cite{hurst2022generation}.
Third, developing and testing financial algorithms requires large volumes of high-quality data. Synthetic data provides an endless supply of training data, enabling thorough backtesting of trading strategies and machine learning models without the risk of overfitting historical data~\cite{peng2021data}. 

Synthetic data generation also transforms the financial services industry by enabling more accurate risk assessments and fraud detection\cite{assefa2020generating}. Synthetic data generation can identify anomalies and potential risks by simulating financial transactions and market behaviors, allowing financial institutions to implement more effective fraud prevention measures and develop more resilient financial strategies. Furthermore, synthetic data generation can support compliance with regulatory requirements by providing detailed, real-time reporting and analysis of financial activities~\cite{wiese2020quant}. In the context of human resources, synthetic data generations can model workforce dynamics, including employee performance, engagement, and turnover. By analyzing these models, businesses can develop strategies to improve employee satisfaction, enhance productivity, and reduce turnover rates. For example, synthetic data generation can simulate the impact of various HR policies on workforce morale and performance, helping HR departments to implement the most effective practices.

\subsection{Other Applications}
The techniques for synthetic data generation described in this paper have far-reaching implications beyond the specific domains covered. Here are some notable applications: 
\begin{itemize}[leftmargin=*]
\item Retail and Marketing: In retail, synthetic data can model customer interactions, purchasing behaviors, and inventory management~\cite{schneider2018flexible}. This aids in developing personalized marketing strategies, optimizing supply chains, and improving customer service without infringing on individual privacy. 
\item Environmental Studies: Synthetic data can simulate environmental conditions, weather patterns, and ecological interactions~\cite{smith2009improving}. This is particularly useful for studying climate change, biodiversity, and conservation efforts, allowing researchers to test hypotheses and model practical scenarios without the constraints of limited real-world data. 
\item Urban Planning and Development: In urban planning, synthetic data can be used to simulate population growth, traffic flows, and infrastructure development~\cite{papyshev2021exploring}. This helps city planners and developers make informed decisions about resource allocation, transportation systems, and sustainable development initiatives. 
\item Software Development and Testing: In software development, synthetic code generation can simulate various coding scenarios, bug patterns, and software behaviors~\cite{li2022competition}. This is particularly useful for testing and debugging, as it allows developers to identify and fix potential issues without the constraints of existing codebases. Synthetic code can also aid in developing personalized coding assistants, optimizing software performance, and improving the reliability of code releases~\cite{ye2024uncovering}. Additionally, by generating diverse and extensive code samples, developers can enhance machine learning models for code completion and error detection, ultimately leading to more efficient and robust software development processes. 
\end{itemize}

\section{Generative AI}
\label{sec:generative_ai}
Generative AI models refer to a wide class of AI methods that could learn the data distribution from existing data objects and generate novel structured data objects, which fall into the category of unsupervised learning. 
Generative AI models, also known as deep generative models, or distribution learning methods, learn the data distribution and samples from the learned distribution to produce novel data objects. 
In this section, we investigate several generative AI models that are frequently used in synthetic data generation, including the language model in Section~\ref{sec:language_model}, 
variational autoencoder (VAE) in Section~\ref{sec:vae}, 
generative adversarial network (GAN) in Section~\ref{sec:gan}, reinforcement learning (RL) in Section~\ref{sec:rl}, 
and diffusion model in Section~\ref{sec:diffusion}. 
Table~\ref{table:generative_ai} compares various generative AI methods from several aspects.

\begin{table*}[ht!]
\centering
\caption{Comparison of all the generative AI methods from different aspects. }
%\resizebox{\columnwidth}{!}{
\scalebox{0.99}{
\begin{tabular}{p{0.50\columnwidth}|p{0.15\columnwidth}|p{0.36\columnwidth}|p{0.08\columnwidth}|p{0.16\columnwidth}|p{0.18\columnwidth}}
\toprule[0.9pt]  
Method & Supervision & NN Architecture & MLE  & With latent variable & Paper \\ 
\midrule  
Language model (LM) & no & autoregressive model & yes & no & \cite{ghosh2017affect} \\  
self-supervised learning (SSL) & no & encoder (representation)  & yes & no &  \\ 
variational autoencoder (VAE) & no & encoder-decoder & yes & yes & \cite{kingma2013auto} \\ 
generative adversarial network (GAN) & no & generator \& discriminator & yes  & yes & \cite{goodfellow2014generative,goodfellow2014distinguishability} \\
diffusion (score-based model) model & no & representation & yes & no & \cite{song2020score} \\ 
reinforcement learning (RL) & yes & policy network or Q-network  & no & no & \cite{sutton2018reinforcement} \\ 
 % & & & \\ 
\bottomrule[0.9pt] 
\end{tabular}
}
\label{table:generative_ai}
\end{table*}

\subsection{Language Model}
\label{sec:language_model} 
The language model was originally designed to model natural language. 
It is able to learn structured knowledge from massive unlabelled sequence data. 
Specifically, suppose the sequence has $N$ tokens, denoted $X = [x_1, \cdots, x_N]$, then the probability distribution of the sequence can be decomposed as the product of a series of conditional probabilities, 
\begin{equation}
p(X) = p\big([x_1, \cdots, x_N]\big) = \prod_{i=1}^{N} p(x_i | x_1, \cdots, x_{i-1}), 
\end{equation}
where a single conditional probability $p(x_i | x_1, \cdots, x_{i-1})$ denote the probability of the token $x_i$ given all the tokens before $x_i$. 
The conditional probability can be modeled by the recurrent neural network (RNN). 
The language model can be used to generate all types of sequence data, such as natural language~\cite{ghosh2017affect}, electronic health records~\cite{lee2018natural}, etc. 
The language model can be combined with other deep learning models, such as variational autoencoder (VAE) and generative adversarial network (GAN), which will be described later.

\subsection{Self-Supervised Learning (SSL)}
\label{sec:ssl} 

Labeled data are expensive to acquire so the number of available labeled data is usually limited. 
To address this issue, self-supervised learning (SSL) was proposed. This learning paradigm curates the supervision signal from the data itself. It is parallel to supervised learning and unsupervised learning. 
Different from supervised learning, self-supervised learning can learn from massive unlabeled data. 
Self-supervised learning is usually used as a pretraining strategy to learn the representation from massive unlabelled data~\cite{hu2019strategies}. 
The core idea of self-supervised learning is to mask a subset of the raw data feature and build a machine learning model to predict the masked data. 
then the pre-trained machine learning model (usually a neural network) is used as a ``warm start’’, and is furtherly finetuned for the downstream applications.

\subsection{Variational Autoencoder (VAE)}
\label{sec:vae}

Variational autoencoder (VAE)~\cite{kingma2013auto} employs a continuous latent variable to characterize the data distribution.
Specifically, it contains two neural network modules: encoder and decoder. 
The objective of the encoder is to convert the data object into a continuous latent variable. 
Then decoder takes the latent variable as the input feature and reconstructs the data object. 

Formally, suppose the data object is denoted $x$, the latent variable is a $d$-dimensional real-valued vector $z$, the encoder is $p(z|x)$, and the decoder is $q(x|z)$. 
The learning objective contains two parts: (1) reconstruct the data object $x$ and (2) encourage the distribution of latent variables to be close to the normal distribution.

The Kullback-Leibler (KL) divergence measures the difference between two probability distributions. Given two probability distributions $p_1(x)$ and $p_2(x)$ on the same continuous domain, KL divergence between them is formally defined as 
% \begin{equation}
% \label{eqn:kl_divergence}
%  \text{KL}(p_1 || p_2) 
% = \int_{x} p_1(x) \log \frac{p_1(x)}{p_2(x)} dx 
% = \int_{x} p_1(x) \big[ \log p_1(x) - \log p_2(x) \big] dx. 
% \end{equation}

\begin{align*}
\label{eqn:kl_divergence}
 \text{KL}(p_1 || p_2)  &= \int_{x} p_1(x) \log \frac{p_1(x)}{p_2(x)} dx  \\
&= \int_{x} p_1(x) \big[ \log p_1(x) - \log p_2(x) \big] dx. \\
\log p(x) & = \log \int_z p(z) p(x|z) dz \\
 & \ge \mathbb{E}_{q(z|x)} \big[ \log p(x|z) \big] - D_{KL}(q(z|x) || p(z)) \\  &\triangleq \text{ELBO}.
\end{align*}

% \begin{equation} 
% \label{eq:variation_autoencoder}
% \begin{aligned}
% \log p(x)
% & = \log \int_z p(z) p(x|z) dz  \ge \mathbb{E}_{q(z|x)} \big[ \log p(x|z) \big] - D_{KL}(q(z|x) || p(z))  \triangleq \text{ELBO}.
% \end{aligned}
% \end{equation}
% \begin{align*}
%  \label{eq:variation_autoencoder}
% \log p(x) & = \log \int_z p(z) p(x|z) dz \\
% & \ge \mathbb{E}_{q(z|x)} \big[ \log p(x|z) \big] - D_{KL}(q(z|x) || p(z))  \triangleq \text{ELBO}.
% \end{align*}
where $p(z)$ is the normal distribution and is used as the prior distribution. 
VAE encourages the distribution of latent variables to be close to normal distribution. 
Then during the inference phase, we sample latent variables from the normal distribution and generate the novel data objects. 
There are several VAE variants, such as disentangled VAE~\cite{burgess2018understanding}, hierarchical VAE~\cite{vahdat2020nvae}, and sequence VAE~\cite{gomez2018automatic}.

\subsection{Generative Adversarial Network (GAN)}
\label{sec:gan}

Generative adversarial network (GAN)~\cite{goodfellow2014generative,goodfellow2020generative,zhang2020dstyle} formulates the generation problem into a supervised learning task. 
Specifically, it comprises two neural network modules: discriminator and generator. 
The objective of the generator is to generate data that are close to the real data, 
By comparison, the objective of the discriminator is to discriminate the fake data (generated by the generator) from the real ones. 
It performs a binary classification task, where the real data from the training set are regarded as the positive samples; the generated data (by generator) are regarded as the negative samples. 
generator and discriminator are trained in a mini-max manner.

Formally, the generator is denoted $G_{}(z)$, and the discriminator predicts a probabilistic score for a data object and is denoted $D(x)$. 
The learning objective is formulated as 
\begin{equation} 
\label{eq:generative_adversarial_net}
\begin{aligned}
 \min_{G} \max_{D} \mathcal{L}(D, G) 
= \ \mathbb{E}_{x \sim \text{training set} } [\log D(x)] \\ + \mathbb{E}_{z \sim p(z)} \big[\log (1-D(G(z)) \big], 
\end{aligned}
\end{equation}
where $z$ is the latent variable and is drawn from the normal distribution $p(z)$ to enhance the diversity of the generated data objects.

When learning GAN, the generator and discriminator are optimized alternatively. 
\begin{itemize}
\item optimize generator and fix discriminator: the objective function becomes 
\begin{equation}
\min \mathcal{L}(G) = \mathbb{E}_{z \sim p(z)} \big[\log (1-D(G(z)) \big], 
\end{equation}
where the generator is optimized to generate data that is close to the real data (with higher discriminator's scores). 
\item optimize discriminator and fix generator: the objective function reduces to a binary classification problem, 
\begin{equation}
\begin{aligned}
\max \mathcal{L}(D) =  \mathbb{E}_{x \sim \text{training set} } [\log D(x)] \\ +  \mathbb{E}_{z \sim p(z)} \big[\log (1-D(G(z)) \big], 
\end{aligned}
\end{equation}
which can be seen as a cross-entropy loss function, where the real data objects from the training set are seen as positive samples while the synthetic data objects $G(z)$ are seen as negative samples. 
\end{itemize}

Then we discuss a popular variant of GAN. 
The Wasserstein Generative Adversarial Network (W-GAN) was proposed in 2017 and aims to enhance the stability of learning, accelerate the training process, and get rid of problems like mode collapse~\cite{arjovsky2017wasserstein}.

\subsection{Reinforcement Learning (RL)}
\label{sec:rl}

Reinforcement learning (RL) focuses on addressing sequential decision-making problems~\cite{zhang2022adapting}. 
It can be used in synthesis data generation by growing a basic component at one time and generating data objects sequentially. 
It formulates sequential decision-making as a Markov decision process (MDP)~\cite{sutton2018reinforcement}. 
Markov decision process assumes that given the current state, the future state of the stochastic process does not depend on the historical states. 
Suppose the state at the time $t$ is $x^t$, Markov decision process satisfies 
\begin{equation}
p(x^{t+1}|x^{t}, x^{t-1}, x^{t-2}, ...) = p(x^{t+1}|x^{t}).     
\end{equation}
At the time $t$, given the state $x^t$, the RL agent would generate an action $a^t$ from action space, which is denoted $p_{\theta}(a^t | s^t)$, $\theta$ is the parameter of the RL agent. 
After performing the action, the system would jump into the next state $x^{t+1}$, i.e., $x^{t+1} = f(x^t, a^t)$. 
At the same time, the system would receive the reward $r(x^t)$ from the environment, where $r(\cdot)$ is called the reward function.  
The goal is to learn an agent that can receive the maximal expected reward in total. 
\begin{equation}
\underset{\theta}{\arg\max}\ \ \mathcal{L}(\theta) =  \sum_{t=1}^{\infty}\mathbb{E}_{p_{\theta}(a^t | x^t)} [r(x^t)]. 
\end{equation}
The objective function is not differentiable with parameter $\theta$. We use policy gradient to obtain an unbiased estimator of the objective gradient $\nabla_{\theta} \mathcal{L}(\theta)$~\cite{sutton2018reinforcement} and then use stochastic optimization methods to maximize the expected reward. 
Generating synthesis data can be viewed as sequential decision-making by sequentially generating one basic structure. 

\subsection{Diffusion Model}
\label{sec:diffusion}

The diffusion model, also known as the score-based model or score matching method, was proposed in recent years~\cite{song2020score} and is widely validated in many generative AI problems such as speech synthesis~\cite{chen2020wavegrad}. 

Specifically, suppose the data object is $x$, and the likelihood function is denoted $p(x)$. 
We are interested in estimating the gradient of the logarithm of the likelihood function.

Diffusion models~\cite{sohl2015deep,ho2020denoising} are inspired by non-equilibrium thermodynamics and can be split into the forward and backward diffusion processes. During the forward diffusion process, diffusion models will gradually add Gaussian noise to the data, and the last-step data will follow an isotropic Gaussian. The reverse diffusion process will revert such a process and construct the data from noise distribution.

More rigorously, we can define the forward process as from the actual data $x_0 \sim p(x)$ to the random noise $x_T$ with $T$ diffusion steps. Let us first assume that for the forward process, the Gaussian distribution is
\begin{equation*}
\begin{aligned}
q(x_t | x_{t-1})
& = \mathcal{N}(x_{t}; \sqrt{1-\beta_t} x_{t-1}, \beta_t I),
\end{aligned}
\end{equation*}
where $\beta_t \in (0, 1)$. Then, the corresponding backward process is 
\begin{equation*} \label{eq:diffusion_backward}
\small{
\begin{aligned}
& p_\theta(x_{t-1}|x_t) \\  
 = & \mathcal{N}(x_{t-1}; \mu_\theta(x_t, t), \Sigma_\theta(x_t,t))\\
 = & \mathcal{N}(x_{t-1}; \frac{1}{\sqrt{\alpha_t}} \big( x_t - \frac{\beta_t}{\sqrt{1 - \bar \alpha_t}} \epsilon \big), \frac{1-\bar \alpha_{t-1}}{1 - \bar \alpha_t} \beta_t),
\end{aligned}
}
\end{equation*}
where $\epsilon \sim \mathcal{N}(0,I)$ follows the standard Gaussian, $\alpha_t = 1-\beta_t$, and $\bar \alpha_t = \prod_{i=1}^t \alpha_i$.

The objective of diffusion models is to estimate the variational lower bound (VLB) of the negative log-likelihood of data distribution:
\begin{equation*}
\begin{aligned}
\log p(x) \ge - \mathbb{E}_{q(x_{1:T}|x_0)} [\log \frac{ q(x_{1:T}|x_0) }{p_\theta(x_{0:T})}] = - \mathcal{L}_{\text{VLB}}.
\end{aligned}
\end{equation*}
The VLB can be rewritten as:
\begin{equation*}
\small{
\begin{aligned}
& \mathcal{L}_{\text{VLB}} 
 = \underbrace{KL[q(x_T|x_0) || p_\theta(x_T)]}_{\mathcal{L}_T} \\
& ~~~ + \sum_{t=2}^T \underbrace{KL[q(x_{t-1}|x_t,x_0) || p_\theta(x_{t-1}|x_t)]}_{\mathcal{L}_{t-1}} \underbrace{- \mathbb{E}_q [\log p_\theta(x_0|x_1)]}_{\mathcal{L}_0}.
\end{aligned}
}
\end{equation*}
Here $\mathcal{L}_T$ is a constant and can be ignored, and diffusion models~\cite{ho2020denoising} have been using a separate model for estimating $\mathcal{L}_0$. For $\{\mathcal{L}_{t-1}\}_{t=2}^T$, we model a neural network to approximate the conditionals during the reverse process, {i.e.,}, we want to train $\mu_\theta(x_t,t)$ to predict $\frac{1}{\sqrt{\alpha_t}} \big( x_t - \frac{\beta_t}{\sqrt{1 - \bar \alpha_t}} \epsilon \big)$. If we plug this into the closed-form solution of the KL-divergence between two multivariate Gaussian distributions, we will have the following for $t = 1, \cdots, T-1$:
\begin{equation*}
\mathcal{L}_t = \mathbb{E}_{x_0,z} \Big[\| \epsilon_t - \epsilon_\theta(\sqrt{\bar \alpha_t}x_0 + \sqrt{1-\bar \alpha_t}\epsilon_t, t) \|^2 \Big].
\end{equation*}
The diffusion model has achieved wide success in many downstream synthetic problems~\cite{liu2021graphebm,fu2021differentiable,weng2021diffusion,somepalli2022diffusion}. 
As a summarization, Table~\ref{table:generative_ai} compares various generative AI methods from several aspects.

\subsection{Multimodal Learning}
\label{sec:multimodal}
Multimodal data refers to datasets that integrate multiple types of data, such as text, images, audio, and numerical values. This type of data provides a comprehensive view by combining different sources of information, which is crucial for tasks requiring a holistic understanding of complex scenarios. In fields like healthcare~\cite{huang2021therapeutics}, finance, and autonomous systems, multimodal data enables more accurate and robust analysis and decision-making by leveraging the strengths of each data type. For instance, in drug discovery, multimodal data can combine genomic data, chemical structures, and clinical outcomes to enhance the prediction of drug efficacy and safety~\cite{fu2022hint,lu2024uncertainty}.

Synthetic multimodal data generation involves creating artificial datasets that integrate multiple types of data, such as text, images, audio, and numerical data, to simulate real-world scenarios. This technique is particularly valuable in fields like healthcare~\cite{choi2017generating}, finance~\cite{assefa2020generating}, and education systems~\cite{bautista2021protecting}, where data is often complex and heterogeneous.

Then, we review some cutting-edge techniques for synthetic multimodal data generation. 
GANs can generate one type of data from another, such as images from textual descriptions or audio from images. This cross-modal generation capability is essential for creating cohesive multimodal datasets~\cite{frid2018synthetic}. 
Recently, ChatGPT~\cite{ventayen2023openai} supports multimodal data generation, including image, text, and numerical features.

\section{Privacy Risks and Prevention}
\label{sec:privacy}

Open release and free data exchange would benefit research and industry development. However, there are cases where datasets exist but cannot be publicly disclosed due to privacy concerns.  
%In the meantime, the emergence and the enforcement of privacy regulations such as HIPAA\footnote{\url{https://www.hhs.gov/hipaa/index.html}}, EU General Data Protection Regulation (GDPR\footnote{\url{https://gdpr-info.eu}}) and the California Consumer Privacy Act (CCPA\footnote{\url{https://oag.ca.gov/privacy/ccpa}}) make it infeasible for sensitive data with privacy implications. 
Regulated data, such as clinical and genomics data in raw form, may not be shared, and one solution is to share synthesized data instead. 

\subsection{Privacy Risks in Data Synthesis}

Due to the utility goal of data synthesis, the synthesized data tends to preserve the distribution of the original data. 
Therefore, the deployment of these models could be subject to privacy leakage. For deep neural network-based approaches, membership inference attack~\cite{shokri2017membership,truex2019demystifying} would identify if an input is in the training data or not and thus can be used to determine how close the synthesized data is to the original data. At the feature level, sensitive attributes such as skin color can be inferred from the behavior of the deep learning model~\cite{melis2019exploiting}, and even the single training instance can be reconstructed~\cite{zhu2019deep,wei2020framework,geiping2020inverting}. For generative AI models, the generative learning process and the high complexity of the model jointly encourage a distribution that is concentrated around training samples.
By repeatedly sampling from the distribution, there is a considerable chance of recovering the training samples or attributes~\cite{hayes2019logan,hitaj2017deep,wang2019beyond,ganev2023inadequacy,hilprecht2019monte,xu2022mace,stadler2022synthetic}, or the membership of the training data~\cite{chen2020gan}.

%record linkage using sensitive attributes~\cite{reiter2009estimating}drechsler2008accounting,

\begin{table*}[ht!]
\centering
\caption{Summarization of privacy prevention strategies in synthetic data generation. }
%\resizebox{\columnwidth}{!}{
\scalebox{0.95}{
\begin{tabular}{p{0.11\columnwidth}|p{0.50\columnwidth}|p{0.50\columnwidth}|p{0.44\columnwidth}|p{0.36\columnwidth} }
\toprule[0.9pt]
Paper & Privacy-enhancing Techniques & Generative AI & DNN &  Data Format \\ 
\hline 
\cite{abay2019privacy} & differential privacy &  autoencoder & autoencoder  & attribute \\ 
\cite{lee2020generating} & differential privacy & VAE + GAN  & recurrent autoencoder & EHR\\
\cite{acs2018differentially} & differential privacy  &  generative  artificial neural networks   & kernel k-means  &  image and text  \\ 
\cite{jordon2018pate} & differential privacy (PATE) & GAN & DNN   & attribute \\ 
\cite{chen2012differentially} & differential privacy & n-gram  & NA (w.o. DNN) & sequential/time series \\
\cite{cunningham2021real} & local differential privacy &n-gram  & NA (w.o. DNN) & trajectory \\
\cite{du2023ldptrace} & local differential privacy & Markov probabilistic model  & NA (w.o. DNN) &  trajectory\\
\cite{he2015dpt} & differential privacy & Markov probabilistic model  & NA (w.o. DNN) & trajectory \\
\cite{wang2017protecting} & differential privacy & Markov probabilistic model  & NA (w.o. DNN) & social media trajectory \\
\cite{gursoy2018differentially} & differential privacy & Markov probabilistic model  & NA (w.o. DNN) & trajectory \\
\cite{mir2013dp} & differential privacy &  distribution estimation & NA (w.o. DNN) & location \\
\cite{roy2016practical} & differential privacy & distribution estimation & NA (w.o. DNN) & trajectory \\
\cite{bindschaedler2016synthesizing} & plausible deniability & Hidden Markov Models  & NA (w.o. DNN) & trajectory \\
\cite{wang2023privtrace} & differential privacy & Markov chain model  & NA (w.o. DNN) & trajectory \\
\cite{narita2024synthesizing} & differential privacy & probabilistical transform  & NA (w.o. DNN) & trajectory \\
\cite{bindschaedler2017plausible} &  plausible deniability & probabilistical transform &  NA (w.o. DNN)   & attribute \\ 
\cite{tseng2020compressive} & compressive privacy  &  GAN & DNN & image \\ 
\cite{zhang2018differentially} & differential privacy & GAN & DNN & image   \\ 
\cite{xie2018differentially} & differential privacy & GAN  &  DNN & image and EHR  \\ 
\cite{xu2019ganobfuscator} &differential privacy  & GAN & DNN & image \\ 
\cite{triastcyn2020federated} & differential privacy & GAN  & DNN &  image \\ 
\cite{lu2017poster} & differential privacy &  GAN & DNN & attribute (tabular) and graph \\ 
\cite{liu2019ppgan} & differential privacy &  GAN & DNN & image and EHR \\ 
\cite{frigerio2019differentially} & differential privacy &  GAN & DNN & time series \\ 
\cite{beaulieu2019privacy} & differential privacy &  GAN & DNN & attribute \\ 
\cite{astolfi2021generating} & differential privacy &  GAN & DNN & attribute (tabular) \\ 
\cite{chen2020gs} & differential privacy &  GAN & DNN & image  \\
\cite{cao2021don} & differential privacy & GAN  & DNN & image \\
% \cite{tantipongpipat2021differentially} & differential privacy & GAN  & autoencoder & attribute \\
\cite{torfi2022differentially} & differential privacy & GAN  & autoencoder & attribute \\
\cite{fan2021dpnet} & differential privacy & GAN  & DNN &  trajectory\\
\cite{long2021g} & differential privacy (PATE) & GAN  & DNN & image \\
\cite{cunningham2022geopointgan} & local differential privacy & GAN &  DNN & spatial point \\
\cite{vinaroz2022hermite} & differential privacy & Maximum Mean Discrepancy   & Hermite polynomial features & attribute (tabular) and image \\
\cite{harder2021dp} & differential privacy & Maximum Mean Discrepancy  & Random feature mean embeddings & image \\
\cite{chen2015differentially} & differential privacy & Markov random field  & NA (w.o. DNN) & attribute \\
\cite{cai2021data} & differential privacy  & Markov random field  & NA (w.o. DNN)  & attribute \\ 
\cite{mckenna2022aim} & differential privacy & Probabilistic graphical models & NA (w.o. DNN) & attribute \\
 \cite{lin2022dpview} &  differential privacy & Maximum Cardinality Matching &  NA (w.o. DNN) & attribute (tabular) \\
\cite{chandrasekaran2021causally} &  differential privacy & Bayesian network &  NA (w.o. DNN) & attribute\\
\cite{zhang2017privbayes} & differential privacy & Bayesian network & NA (w.o. DNN) & attribute \\ 
\cite{ge2020kamino} & differential privacy & statistical database  & NA (w.o. DNN) & attribute \\ 
\cite{gaboardi2014dual} & differential privacy & statistical queries &  A (w.o. DNN) & attribute \\ 
\cite{hardt2012simple} & differential privacy & statistical queries &  A (w.o. DNN) & attribute \\ 
\cite{zhang2021privsyn} & differential privacy & Graduate Update Method & NA (w.o. DNN) & attribute   \\ 
\cite{chen2018differentially} & differential privacy &  autoencoder & autoencoder & image and attribute \\
\cite{ bao2021synthetic} &  differential privacy & autoencoder &  autoencoder & text/image QA and attribute\\
% \cite{bo2019er} & differential privacy &  autoencoder & GRU/LSTM & text \\
\cite{liu2022privacy} & data replacement and item regularizer & latent space projection  & MLP & attributes  \\ 
\cite{chen2022dpgen} & differential privacy &  Langevin Markov chain Monte Carlo & Energy-based Model & image  \\
\cite{zhang2018calm} & local differential privacy  & Maximum Entropy estimation & NA (w.o. DNN) & attribute \\ 
\cite{qardaji2014priview} &  differential privacy  & Maximum Entropy estimation  & NA (w.o. DNN) & attribute \\ 
%%% you may add this paper Synthetic data augmentation using GAN for improved liver lesion classification.
\bottomrule[0.9pt]
\end{tabular}
}
\label{table:summarize_privacy}
\end{table*}

\subsection{Privacy Protection in Data Synthesis}

Solutions have been proposed in two broad categories. In the first category, 
different data anonymization-based approaches such as K-anonymity~\cite{sweeney2002k,samarati1998generalizing,samarati2001protecting} and nearest marginal~\cite{barak2007privacy} to sanitize data so that it cannot be easily re-identified. These data anonymization approaches involve replacing sensitive data with fictitious yet realistic data. It is often used to protect the data while maintaining its usability for testing or development purposes.
However, they often do not provide rigorous privacy guarantees~\cite{abay2019privacy}. 
%ye2020alpha, qian2022co, ding2011differentially
In the second category, synthetic data generation approaches have been proposed to generate realistic synthetic data using rigorous differential privacy definitions~\cite{bindschaedler2017plausible,dwork2014algorithmic,duchi2013local} for various applications. These approaches involves adding noise to the data to prevent the identification of individuals in the dataset while preserving the statistical properties of the data. This is particularly useful in scenarios where data needs to be shared but individual privacy must be maintained. In particular, Bindschaedler et al.~\cite{bindschaedler2017plausible} introduced the idea of plausible deniability instead of directly adding noise to the generative model. This mechanism results in input indistinguishability that
means by observing the output set (i.e., synthetics) an adversary cannot make sure whether a particular data record was
in the input set (i.e., real data). With the help of generative modeling,  Acs et al.~\cite{acs2018differentially} clusters the original datasets into k clusters with differentially private kernel k-means and produce synthetic data for each cluster. 
By comparison, Liu et al~\cite{liu2022privacy} introduce two-level privacy-preserving synthetic data generation. At the data level, a selection module is used to select the items which contribute less to the user’s preference. At the item level, a synthetic item generation module is developed to create the corresponding synthetic item.

 %tantipongpipat2021differentially,
 Taking advantage of the GAN, several methods are proposed
to generate synthetic data to get better effect~\cite{zhang2018differentially,beaulieu2019privacy,jordon2018pate,xie2018differentially,lee2020generating,torfi2022differentially,fan2021dpnet} which closely matches the distribution of the
source data than the hidden Markov model-based approach~\cite{dahmen2019synsys}, RBF based approach~\cite{cai2021data}, Bayesian network-based~\cite{bao2021synthetic}, and Auto-encoder based approach~\cite{abay2019privacy}. 
Xie et al~\cite{xie2018differentially} propose DPGAN by adding noise on the gradient of the Wasserstein distance with respect to the training data. This approach does not adopt the optimization strategy to improve the training stability and convergence speed. To address these problems, Zhang et al.~\cite{zhang2018differentially} proposed dp-GAN, a general private data
publishing framework for rich semantic data without the requirement of tag information
compared to~\cite{beaulieu2019privacy}. By comparison, 
Beaulieu-Jones et al.~\cite{beaulieu2019privacy}
trained the discriminator under differentially private SGD,
which generates plausible individuals of clinical datasets. 
Tseng and Wu~\cite{tseng2020compressive} apply compressive privacy~\cite{kung2017compressive} for CPGAN, which would generate compressing representations that retain high utility. Jordon et al. ~\cite{jordon2018pate} modifies the Private Aggregation of Teacher Ensembles (PATE) framework and applies it to the discriminator of GANs. The proposed approach perceives the discriminator as a classifier and utilizes its output as knowledge such that the student learns from noisy labels that are obtained through privately aggregating the discriminator’ votes. This allows a tight bound on the influence of any individual sample on the model, resulting in tight differential privacy guarantees and thus an improved performance over models for data synthesis.  By comparison, Long et at.~\cite{long2021g} applies teacher-student-based differential privacy to the generator.
While most of these approaches inject noise into the energy function, a differentially private GAN called GANobfuscator~\cite{xu2019ganobfuscator} achieve differential privacy by adding noise within the training procedure.

While centralized differential privacy assumes data aggregators are reliable, local differential privacy (LDP)~\cite{duchi2013local} assumes that aggregators cannot be trusted and relies on data providers to perturb their own data and is used to generate private synthetic datasets that is similar to the private dataset. \cite{zhang2018calm} is inspired by PriView~\cite{qardaji2014priview} but for computing any k-way marginals under the LDP setting for the marginal table release problem. Furthermore, \cite{cunningham2022geopointgan} considers DP at label-level on GAN for synthetic spatial point generation. Apart from LDP in distributed setting, Triastcyn and Faltings~\cite{triastcyn2020federated} propose federated generative privacy that utilizes insufficient local data from multiple clients to train a GAN. The method shares only generators that do not come directly into contact with data and the discriminator remain private. This model can output artificial data, not belonging to any real user in particular, but coming from the common cross-user data distribution.

These privacy-preserving data synthesis methods mainly aim at structured data like tables, which cannot be applied to high dimensionality and complexity. To solve this problem, PriView~\cite{qardaji2014priview} constructs the private k-way marginal tables for $k \geq 3$ by first extracting low-dimensional marginal views from the flat data and adding noise to the views and then applying a reprocessing technique to ensure the consistency of the noisy views. \cite{li2014differentially,patki2016synthetic,gambs2021growing,asghar2019differentially} leverage copula functions for multi-dimensional differentially private synthesization.  Zhang et al.~\cite{zhang2017privbayes} consider repetitive perturbation of the original data as a substitute to the original data with a synthetic data generation technique called PrivBayes. PrivBayes decomposes high dimensional data into low dimensional marginals by constructing a Bayesian network and injects noise into these learned low dimensional marginals to ensure differential privacy and the synthetic data is inferred from these noised marginals. Instead of the Bayesian network, differentially private auto-encoder~\cite{abay2019privacy} significantly improves the effectiveness of differentially private synthetic data release. 
\cite{ge2020kamino} applies data cleaning method~\cite{rekatsinas2017holoclean} to fix the violations on the structure of the data in the synthetic data. 
Instead of using graphical models as the summarization/representation of a dataset~\cite{abay2019privacy,bindschaedler2017plausible,gaboardi2014dual,hardt2012simple,zhang2018differentially},~\cite{zhang2021privsyn} proposes to use a set of large number of low-degree marginals to represent a dataset. The advantage of this approach is that it makes weak assumptions about the conditional independence among attributes, and simply tries to capture correlation relationships that are in the dataset. Meanwhile, the method is especially attractive under differential privacy for its straightforward sensitivity measurement, reduced noise variance, and efficient privacy cost. \cite{vinaroz2022hermite} leverages the Hermite polynomial features to 
encapsulate a higher degree of information within a smaller order of feature. \cite{chen2015differentially} constructs a graph that explore pairwise dependence between attributes and applies the junction tree algorithm to obtain the Markov random field (MRF), from which the noisy marginals are generated and the synthetic data are sampled. 
%beaulieu2016semi,

While private synthetic data generation algorithms are agnostic to downstream tasks, it is important to meet the utility requirements for downstream use.~\cite{wang2024post} proposes post-processing via resampling from the synthetic data to filter out samples that do not meet the selected utility measures, thus improving the utility of synthetic data.

\subsection{Privacy Threats in Foundation Models}

Entering the era of foundation models, recent research has demonstrated that training data can be exposed from large language models~\cite{carlini2021extracting} as well as stable diffusion~\cite{carlini2023extracting}. In both types of models, attackers can generate sequences from the trained model and identify those memorized from the training set.
Studies have shown that a sequence that appears multiple times in the training data is more likely to be generated than a sequence that occurred only once~\cite{meehan2020non,feng2021gans,somepalli2022diffusion}. Accordingly, 
Kandpal et al~\cite{kandpal2022deduplicating} propose to deduplicate the training data that appears multiple
times such that the privacy risks in language models is mitigated. \cite{dockhorn2022differentially} is the first work to enforce privacy using differentially private stochastic gradient descent (DP-SGD) in diffusion models. Several attempts has been made to 
reduces the noise in the gradient during DP-SGD training and improves the generative quality in diffusion models, via semantic-aware pretraining~\cite{tsai2024differentially,wang2024dp}, latent information~\cite{lyu2023differentially}, and retrieval-augmented generation~\cite{lebensold2024dp}. In the meantime, differential privacy has been heavily invested in privacy protection of large language models~\cite{yu2021differentially}.

Given that we are still at the very early stage of the generative foundational models, the potential of the foundation models for data synthesis has not been fully explored. While more possible privacy threats on the foundation models are yet to be discovered, existing privacy measures may be inadequate to meet its demands of privacy. Further investigation is needed to 
design countermeasures that would mitigate the memorization and generalization problems for privacy protection.

\section{Fairness}
\label{sec:fairness}

Generating synthetic data that reflect the important underlying statistical properties of the real-world data may also inherit the bias from data preprocessing, collection, and algorithms~\cite{wei2024trustworthy}. Minority groups can often end up being under-represented in synthetic data~\cite{oprisanu2021measuring,pereira2106analysis,ganev2022robin}.
The fairness problem is currently addressed by three types of methods~\cite{mehrabi2021survey}: (i) preprocessing, which revises input data to remove information correlated to sensitive
attributes, usually via techniques like massaging, reweighting, and sampling. (ii) in-processing, which adds fairness constraints to the model learning process; and (iii) post-processing, which adjusts model predictions after the model is trained.

Most existing fairness-aware data synthesis methods leverage preprocessing techniques. The use of balanced synthetic datasets created by GANs to augment classification training has demonstrated the benefits for reducing disparate impact due to minoritized subgroup imbalance~\cite{abusitta2019generative,tanaka2019data,mariani2018bagan}.~\cite{barbierato2022methodology} models bias using a probabilistic network exploiting structural equation modeling as the preprocessing to generate a fairness-aware synthetic dataset. Authors in~\cite{xu2018fairgan} leverage GAN as the pre-processing for fair data generation that ensures the generated data is discrimination free while maintaining high data utility. By comparison, \cite{sattigeri2019fairness} is geared towards high dimensional image data and proposes a novel auxiliary classifier GAN that strives for demographic parity or equality of opportunity. 
However, preprocessing would require the synthesized data provider to know all correlations, biases, and distributions of variables in the existing datasets as a priori. 
Compared to preprocessing, the latter two categories are less-developed for fair data synthesis. \cite{van2021decaf} insert a structural causal model in the input layers of the generator, allowing each variable to be reconstructed conditioned on its causal parents for inference time debiasing.

In the meantime, differential privacy amplifies the fairness issues in the original data~\cite{bagdasaryan2019differential}.
\cite{cheng2021can} demonstrate that differential privacy does not introduce unfairness into the data generation process or to standard group fairness measures in the downstream classification models, but does unfairly increase the influence of majority subgroups. Differential privacy also significantly reduces the quality of the images generated from the GANs, decreasing the synthetic data's utility in downstream tasks. To measure the fairness in synthesized data,~\cite{bhanot2021problem} develops two covariate-level disparity fairness metrics for synthetic data. The authors analyze all subgroups defined by protected attributes to analyze the bias.

In the emerging AIGC using foundation models, the generated images and texts may also inherit 
the stereotypes, exclusion and marginalization of certain groups and toxic and offensive information in the real-world data. This  would lead to discrimination and harm to certain social groups. The misuse of such data synthesis approaches by misinformation and manipulation would lead to further negative social impact~\cite{weidinger2021ethical}. 
Given that the quality of the data generated by foundation models is
inextricably linked to the quality of the training corpora, it is essential to regulate the 
real-world data being used to form the data synthesis distribution. While reducing bias in data is important,  the remaining bias in the data may also be amplified by the models~\cite{mehrabi2021survey} or the privacy-enhancing components~\cite{bagdasaryan2019differential}. 
With frequent inspection and sensitive and toxic information removal on both data and model, it will help govern the information generated from those foundation models and ensure the models would do no harm.

\section{Evaluation Strategy}
\label{sec:evaluation}

In this section, we discuss various approaches to evaluating the quality of synthesized data, which is essential for determining the effectiveness and applicability of synthetic data generation methods in real-world scenarios. We categorize these evaluation strategies as follows:
\begin{enumerate}[leftmargin=*]
\item \textbf{Human evaluation}. This method is the most direct way to assess the quality of synthesized data. Human evaluation involves soliciting opinions from domain experts or non-expert users to judge the synthesized data's quality, similarity to real data, or usability in specific applications. 
For example, in speech synthesis, the human evaluator rates the synthesized speech and real human speech in a blind manner~\cite{anumanchipalli2019speech,donahue2018adversarial}. 
However, human evaluation has several drawbacks, including being expensive, time-consuming, error-prone, and not scalable. Additionally, it struggles with high-dimensional data that cannot be easily visualized and evaluated by humans. 
%% tianfan: need some citation 

\item \textbf{Statistical difference evaluation}. This strategy involves calculating various statistical metrics on both the synthesized and real datasets and comparing the results. For example, ~\cite{yi2018enhance,yan2022multifaceted} use first-moment statistics of individual features (e.g., medical concept frequency/correlation, patient-level clinical feature) to evaluate the quality of generated electronic health record (EHR) data. The smaller the differences between the statistical properties of synthetic and real data, the better the quality of the synthesized data.

\item \textbf{Evaluation using a pre-trained machine learning model}. As mentioned in Section~\ref{sec:gan}, in the generative adversarial network (GAN), the discriminator differentiates fake data (synthesized data) from real ones. Consequently, the output of the discriminator can measure how closely synthetic data resembles real data. The performance of the discriminator on the synthesized data can be used as an indicator of how well the generator produces realistic data. This strategy can be applied not only to GANs but also to other generative models where a pre-trained machine learning model is used for evaluation.

\item \textbf{Training on synthetic dataset and testing on the real dataset (TSTR)}. This strategy involves using synthetic data to train machine learning models and assessing their prediction performance on real test data in downstream applications. High performance on real test data indicates that the synthetic data has successfully captured essential characteristics of the real data, making it a useful proxy for training. For example, \cite{esteban2017real} employs synthetic data to train machine learning models and assess their prediction performance on real test data in downstream applications. TSTR can provide insights into the effectiveness of synthetic data for training machine learning models in a wide range of tasks and domains.

\item \textbf{Application-specific evaluation}. Depending on the specific use case or domain, tailored evaluation methods may be employed to assess the quality of synthesized data. These evaluation methods can consider the unique requirements or constraints of the application, such as regulatory compliance, privacy concerns, or specific performance metrics. By evaluating the synthesized data in the context of its intended use, a more accurate assessment of its quality and applicability can be obtained.
\end{enumerate}

These evaluation strategies offer various ways to gauge the quality of synthesized data, helping researchers and practitioners determine the effectiveness of synthetic data generation methods and their applicability in real-world scenarios. Employing a combination of these strategies can provide a more comprehensive understanding of the strengths and weaknesses of the synthesized data, facilitating further improvements in synthetic data generation techniques~\cite{zhao2022meta}.

\section{Challenges and Opportunities}
\label{sec:challenge}

The aim of this research is to present a comprehensive survey of synthetic data generation—a promising and emerging technique in contemporary deep learning. This survey outlines current real-world applications and identifies potential avenues for future research in this field. The utilization of synthetic data has been proven effective across a diverse array of tasks and domains~\cite{nikolenko2021synthetic}. In this section, we delve into the challenges and opportunities presented by this rapidly evolving area.

First and foremost, evaluation metrics for synthetic data are essential to determine the reasonableness of the generated data. In industries like healthcare, where data quality is of paramount importance, clinical quality measures and evaluation metrics are not always readily available for synthetic data. Clinicians often struggle to interpret existing criteria such as probability likelihood and divergence scores when assessing generative models~\cite{chen2021synthetic}. Concurrently, there is a pressing need to develop and adopt specific regulations for the use of synthetic data in medicine and healthcare, ensuring that the generated data meets the required quality standards while minimizing potential risks.

Secondly, due to limited attention and the challenges associated with covering various domains using synthetic data, current methods might not account for all outliers and corner cases present in the original data. Investigating outliers and regular instances and their impact on the parameterization of existing methods could be a valuable research direction~\cite{huang2013rank}. To enhance future detection methods, it may be beneficial to examine the gap between the performance of detection methods and a well-designed evaluation matrix, which could provide insights into areas that require improvement.

Thirdly, synthetic data generation may involve underlying models with inherent biases, which might not be immediately evident~\cite{bhanot2021problem}. Factors such as sample selection biases and class imbalances can contribute to these issues. Typically, algorithms trained with biases in sample selection may underperform when deployed in settings that deviate significantly from the conditions in which the data was collected~\cite{chen2021synthetic}. Thus, it is crucial to develop methods and strategies that address these biases, ensuring that synthetic data generation leads to more accurate and reliable results across diverse applications and domains.

Last but not the least, the rise of foundation models in data synthesis presents both significant challenges and opportunities. On one hand, foundation models can be exploited by malicious actors to create sophisticate jailbreak attacks, deepfakes, discrimination, exclusion and
toxicity problems, misinformation harms, sensitive information disclosure, and malicious use. These models can generate human-like text and realistic images or videos, making it difficult for traditional security measures to detect malicious content. Furthermore, the accessibility and rapid advancement of these technologies lower the barrier for cybercriminals, enabling more sophisticated and widespread attacks. The ability to generate vast amounts of realistic, yet fake, data can also overwhelm and deceive traditional detection systems, leading to an increase in false negatives and undetected breaches. On the other hand, foundation models offer promising opportunities to bolster cybersecurity defenses. AI-driven anomaly detection systems can leverage generative models to simulate various attack scenarios, improving their ability to recognize and mitigate real-world threats. In the meantime, the quest for transparency and interpretability in generative models promotes research into explainable AI.  By proactively addressing these machine learning risks, synthetic data generation can evolve to deliver more ethical, secure, and transparent solutions, ultimately harnessing its full potential to benefit society while mitigating its associated risks.

In general, the use of synthetic data is becoming a viable alternative to training models with real data due to advances in simulations and generative models. However, a number of open challenges need to be overcome to achieve high performance. These include the lack of standard tools, the difference between synthetic and real data, and how much machine learning algorithms can do to exploit imperfect synthetic data effectively. 
Though this emerging approach is not perfect now, with models, metrics, and technologies maturing, we believe synthetic data generation will make a bigger impact in the future.

\section{Conclusion}
In conclusion, machine learning has revolutionized various industries by enabling intelligent computer systems to autonomously tackle tasks, manage and analyze massive volumes of data. However, it still faces several challenges, including data quality, data scarcity, and data governance. These challenges can be addressed through synthetic data generation, which involves the artificial annotation of information generated by computer algorithms or simulations. Synthetic data has been extensively utilized in various sectors due to its ability to bridge gaps, especially when real data is either unavailable or must be kept private due to privacy or compliance risks.

This paper has provided a high-level overview of several state-of-the-art approaches currently being investigated by machine learning researchers for synthetic data generation. We have explored different real-world application domains, and examined a diverse array of deep neural network architectures and deep generative models dedicated to generating high-quality synthetic data. 

To sum up, synthetic data generation has enormous potential for unlocking the full potential of machine learning and its impact on various industries. While challenges persist in the development and application of machine learning technology, synthetic data generation provides a promising solution that can help address these obstacles. Future research can further enhance the functionality of synthetic data generation.

\vspace{0.6cm}

\bibliographystyle{IEEEtran}
%\bibliography{sample-base}
\bibliography{ref}

\vfill

\end{document}